# EARLY DETECTION OF BREAST CANCER USING SVM CLASSIFIER TECHNIQUE

Y.Ireaneus Anna Rejani[+], Dr.S.Thamarai Selvi[*]

[+] Noorul Islam College of Engineering, Tamilnadu, India.
*Professor&Head, Department of Information and technology, MIT, Chennai, Tamilnadu, India.

*Abstract:* This paper presents a tumor detection algorithm from mammogram. The proposed system focuses on the solution of two problems. One is how to detect tumors as suspicious regions with a very weak contrast to their background and another is how to extract features which categorize tumors. The tumor detection method follows the scheme of (a) mammogram enhancement. (b) The segmentation of the tumor area. (c) The extraction of features from the segmented tumor area. (d) The use of SVM classifier.

The enhancement can be defined as conversion of the image quality to a better and more understandable level. The mammogram enhancement procedure includes filtering, top hat operation, DWT. Then the contrast stretching is used to increase the contrast of the image. The segmentation of mammogram images has been playing an important role to improve the detection and diagnosis of breast cancer. The most common segmentation method used is thresholding. The features are extracted from the segmented breast area. Next stage include, which classifies the regions using the SVM classifier. The method was tested on 75 mammographic images, from the mini-MIAS database. The methodology achieved a sensitivity of 88.75%.

## I. 1. INTRODUCTION

Breast cancer is the most common non skin malignancy in women and the second leading cause of female cancer mortality [1]. Breast tumors and masses usually appear in the form of dense regions in mammograms. A typical benign mass has a round, smooth and well circumscribed boundary; on the other hand, a malignant tumor usually has a speculated, rough, and blurry boundary [2], [3].

Computer aided detection (CAD) systems in screening mammography serve as a second opinion for radiologists by identifying regions with high suspicious of malignancy [4]. The ultimate goal of CAD is to indicate such locations with great accuracy and reliability. Thus far, most studies support the fact that CAD technology has a positive impact on early breast cancer detection [5], [6].

There is extensive literature on the development and evaluation of CAD systems in mammography. Most of the proposed system follows a hierarchical approach. Initially the CAD system prescreens a mammogram to detect suspicious regions in the breast parenchyma that serve as candidate location for further analysis. In this the first stage is an algorithm of Gaussian smoothing filter, top hat operation for image enhancement in which the combined operations are applied to the original gray tone image and the higher sensitive lesion site selection of the enhanced images are observed. Then the second stage develops a thresholding method for segmenting tumor area.

SVM is a learning machine used as a tool for data classification, function approximation, etc, due to its generalization ability and has found success in many applications [7-11]. Feature of SVM is that it minimizes and upper bound of generalization error through maximizing the margin between separating hyper plane and dataset. SVM has an extra advantage of automatic model selection in the sense that both the optimal number and locations of the basis functions are automatically obtained during training. The performance of SVM largely depends on the kernel [12], [13].

## II. 2. METHODS

Detection of tumors in mammogram is divided into three main stages. The first step involves an enhancement procedure, image enhancement techniques are used to improve an image, where to increase the signal to noise ratio and to make certain features easier to see by modifying the colors or intensities. Then the intensity adjustment is an image's intensity values to a new range. After the mammogram enhancement segment the tumor area. Then the features are extracted from the segmented mammogram. Then the next stage involves the classification using SVM classifier.

*2.1 Image enhancement:*

Image enhancement can be defined as conversion of the image quality to a better and more understandable level. The enhancement procedure is (a) the mammogram images are filtered by Gaussian smoothing filter based on standard deviation. (b) Perform morphological top hat filtering on the gray scale input image using the structuring element. The top hat filtering is used to correct uneven illumination when the background is dark. The top hat filtering with a dark shaped structuring element to remove the uneven background illumination from an image. (c) The top hat output is decomposed into two scales using discrete wavelet transform and then the image is reconstructed.

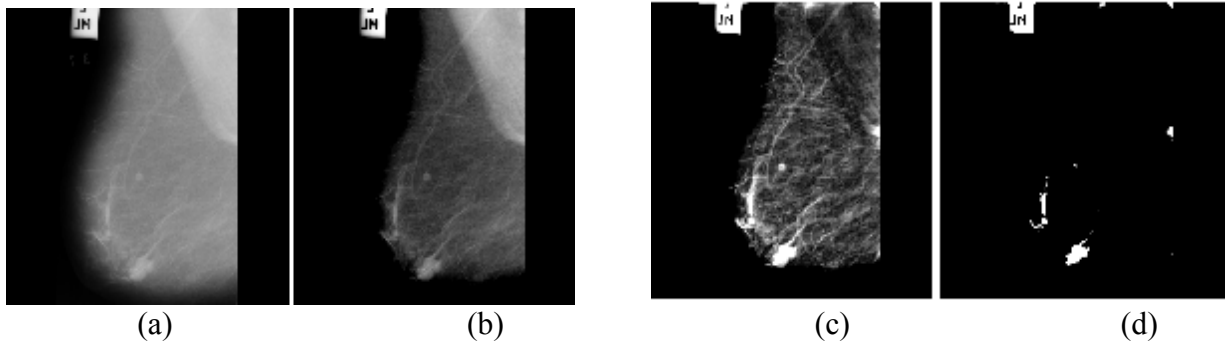

(a)     (b)     (c)     (d)

Figure 1.    (a) Original mammogram. (b) Filtered image. (c) Second level DWT reconstructed mammogram. (d) Tumor segmented output.

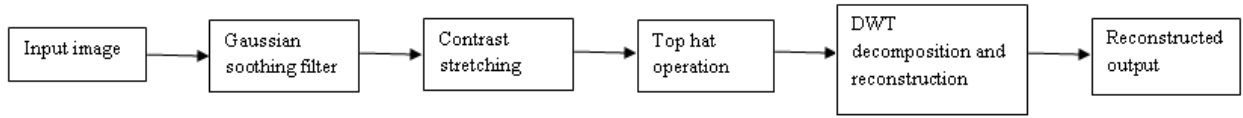

Figure 2.    Block diagram of image enhancement

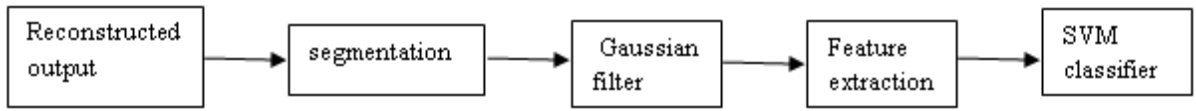

Figure 3.    Block diagram of the proposed method.

*2.2. Segmentation and Feature extraction:*

The enhanced mammogram images are converted to binary images through thresholding at different values. The segmented images are filtered again with Gaussian smoothing filter to eliminate noise. The thresholding is an important step to improve the detection of breast cancer segmentation subdivides an image into its constituent regions. The feature extraction is used to measure the properties from the segmented image are, 1. Area, 2. Centriod, 3.major axis length, 4.minor axis length, 5.eccentricity, 6.orientation, 7.filled area, 8.extrema, 9.solidity, 10.equivdiameter. The area is the scalar value; it computes the actual number of pixels in the region. Then the centroid is the vector and it computes the centre of the tumor region.

*2.3. SVMclassifier:*

Consider the pattern classifier, which uses a hyper plane to separate two classes of patterns based on given examples $\{x(i), y(i)\}_{i=1}^{n}$. Where (i) is a vector in the input space $I=R^k$ and y (i) denotes the class index taking value 1 or 0. A support vector machine is a machine learning method that classifies binary classes by finding and using a class boundary the hyper plane maximizing the margin in the given training data. The training data samples along the hyper planes near the class boundary are called support vectors, and the margin is the distance between the support vectors and the class boundary hyperplanes. The SVM are based on the concept of decision planes that define decision boundaries. A decision plane is one that separates between assets of objects having different class memberships. SVM is a useful technique for data classification. A classification task usually involves with training and testing data which consists of some data instances. Each instance in the training set contains one "target value" (class labels) and several "attributes" (features).

Given a training set of instance label pairs $(x_i,y_i), i=1,\ldots,l$ where $x_i \in R^{un}$ and $y \in (1,-1)^l$, the SVM require the solution of the following optimization problem.

$$\text{Min}_{w, b, \varepsilon} 1/2 w^T w + c \sum_{i=1}^{l} \varepsilon_I$$

Subject to $y_i(w^T \emptyset(x_i) + b) \geq 1 - \varepsilon_I,$

$\varepsilon_i \geq 0.$

Here training vectors $x_i$ are mapped into a higher dimensional space by the function $\emptyset$. Then SVM finds a linear separating hyper plane with the maximal margin in this higher dimensional space>0 is a penalty parameter of the error term. Furthermore, k $(x_i,x_j) = \emptyset(x_i) \emptyset(x_j)$ is called the kernel functions.

There are number of kernels that can be used in SVM models. These include linear polynomial, RBF and sigmoid

$\emptyset = \{x_i * x_j \quad$ linear

$(\gamma x_i x_j + coeff)^d \quad$ polynomial

$\text{Exp}(-\gamma|x_i - x_j|^2) \quad$ RBF

$\text{Tanh}(\gamma x_i x_j + coeff)$ sigmoid$\}$

The RBF is by for the most popular choice of kernel types used in SVM.There is a close relationship between SVMs and the Radial Basis Function (RBF) classifiers. In the field of medical imaging the relevant application of SVMs is in breast cancer diagnosis. The SVM is the maximum margin hyper plane that lies in some space. The original SVM is a linear classifier. For SVMs, using the kernel trick makes the maximum margin hyper plane fit in a feature space. The feature space is a non linear map from the original input space, usually of much higher dimensionality than the original input space. In this way, non linear SVMs can be created. Support vector machines are an innovative approach to constructing learning machines that minimize the generalization error. They are constructed by locating a set of planes that separate two or more classes of data. By construction of these planes, the SVM discovers the boundaries between the input classes; the elements of the input data that define these boundaries are called support vectors.

For Gaussian radial basis function:

$K(x, x') = \exp(-|x-x'|^2 / (2\sigma^2)).$

The kernel is then modified in data dependent way by using the obtained support vectors. The modified kernel is used to get the final classifier.

### III. 3. RESULT

In this paper, the proposed method includes the mammogram image was filtered with Gaussian filter based on standard deviation and matrix dimensions such as rows and columns. Then the filtered image is used for contrast stretching. Then the background of the image is eliminated using top hat operation. Then the top hat output is decomposed into two scales and then use DWT reconstruction. The reconstructed image is used for segmentation. Thresholding method is used for segmentation and then the features are extracted from the segmented tumor area. Then the final stage is classification using SVM classifier.

### IV. 4. CONCLUSION:

To summarize the developed method, the initial step, based on gray level information of image enhancement and segments the breast tumor. For each tumor region extract, morphological features are extracted to categorize the breast tumor. Finally the SVM classifier is used for classification.